\documentclass[sigconf]{acmart}

\AtBeginDocument{%
  \providecommand\BibTeX{{%
    \normalfont B\kern-0.5em{\scshape i\kern-0.25em b}\kern-0.8em\TeX}}}

\setcopyright{acmcopyright}
\copyrightyear{2021}
\acmYear{2021}

\usepackage{ltablex,booktabs,lipsum}
\usepackage{enumitem}
\usepackage{adjustbox}
\usepackage[justification=centering]{caption}
\usepackage{graphicx}
\usepackage{array,tabularx}
\usepackage{multirow} 
\usepackage{enotez}
\setlist[itemize]{noitemsep, topsep=0.5pt}

\usepackage{endnotes}

\let\footnote=\endnote
\begin{document}

\title{Beyond modeling: NLP Pipeline for efficient environmental policy analysis}

\author{Jordi Planas}
\email{jordi@madeingame.cat}
\affiliation{%
  \institution{Omdena}
  \country{Catalonia}
}

\author{Daniel Firebanks-Quevedo}
\email{dafirebanks@gmail.com}
\affiliation{%
  \institution{Data Science for Social Good}
  \country{USA}
}

\author{Galina Naydenova}
\email{galiusha@hotmail.com}
\affiliation{%
  \institution{Omdena}
  \country{Japan}
}

\author{Ramansh Sharma}
\email{sharmar@bxscience.edu}
\affiliation{%
  \institution{Omdena}
  \country{India}
}

\author{Cristina Taylor}
\email{cristina.taylor@msn.com}
\affiliation{%
  \institution{World Resources Institute}
  \country{USA}
}

\author{Kathleen Buckingham}
\email{kathleen@tentree.com}
\affiliation{%
  \institution{veritree}
  \country{USA}
}

\author{Rong Fang}
\email{rong.fang@wri.org}
\affiliation{%
  \institution{World Resources Institute}
  \country{USA}
}

\renewcommand{\shortauthors}{Planas, et al.}

\begin{abstract}
  As we enter the UN Decade on Ecosystem Restoration, creating effective incentive structures for forest and landscape restoration has never been more critical. Policy analysis is necessary for policymakers to understand the actors and rules involved in restoration in order to shift economic and financial incentives to the right places. Classical policy analysis is resource-intensive and complex, lacks comprehensive central information sources, and is prone to overlapping jurisdictions. We propose a Knowledge Management Framework based on Natural Language Processing (NLP) techniques that would tackle these challenges and automate repetitive tasks, reducing the policy analysis process from weeks to minutes. Our framework was designed in collaboration with policy analysis experts and made to be platform-, language- and policy-agnostic. In this paper, we describe the design of the NLP pipeline, review the state-of-the-art methods for each of its components, and discuss the challenges that rise when building a framework oriented towards policy analysis.
  
\end{abstract}



\maketitle

\section{Introduction}
The 2030 Agenda for Sustainable Development was launched by the UN General Assembly in 2015 to address global problems like poverty and environmental degradation\cite{UN2015}. The corresponding actions proposed by the Agenda—17 Sustainable Development Goals (SDGs)—demand nothing short of the transformation of the financial, economic, and political systems that govern our societies. However, as member states recognized at the SDG Summit in 2020, global efforts have thus far been insufficient to deliver adequate change, jeopardizing the Agenda’s promise to current and future generations\cite{UNDESA2020}.

In order to reach the environment-focused SDGs\footnote{Goal 13: Climate Action, Goal 14: Life Below Water, Goal 15: Life on Land}, robust and transparent policy must be put in place. For this to be successful, there are two main issues that need to be addressed: the flow of information between government and citizens\cite{Falla_2018} and policy analysis at scale. This paper explores how we can ensure that policy analysis is done in a reliable, reproducible, and rapid manner.

\subsection{Current challenges in policy analysis}
\label{policyanalysischallenges}
‘Policy’ is a nebulous term that comprises many concepts. In general, ‘public policy’ refers to the ways in which governments seek to influence behavior in order to achieve an objective. Many elements of policy are grouped under legal expressions such as ‘acts’ and ‘laws’, which refer to the implementation format that a policy takes\cite{Kraft2017}. For the purposes of this paper, we use ‘policy’ to refer to the mechanisms that support or enforce a government's objectives. Policy analysis, then, is the evaluation of the design and implementation of policies. Currently, there are three major challenges associated with policy analysis: 

\subsubsection{Time- and resource-intensity}
\label{challenge1}
Governments rarely have employees whose sole responsibility is policy analysis, and even when these employees exist, they often lack the time to adequately conduct and update their analyses\cite{Vining2015}. A common response to this issue in Latin America, for example, is the outsourcing of policy analysis to consultants with the notable exception of Chile, which employs policy analysts in its Office of Agrarian Studies and Policies. However, even with this office dedicated to policy analysis, reports on agricultural programs lag behind implementation by an average of 3 years\cite{ODEPA}. An alternative to overcome this challenge that has been utilized routinely in health policy is the reliance on commissioned rapid reviews, which provide policymakers with relevant summaries of research and take an average of 5–12 weeks to complete\cite{Tricco2017}. The prevalence of rapid reviews in the healthcare sector demonstrates the utility of timely research—a commodity that the restoration policy world currently lacks.  

\subsubsection{Information accessibility}
\label{challenge2}
The environmental policy field lacks central sources that allow policy analysts to search for relevant policy information. While there are some environmental policy databases in existence, each has disadvantages. For example, ECOLEX —operated by FAO, UNEP, and IUCN— is a database of global environmental legislation, but it lacks the most up-to-date documents.\footnote{\href{https://www.ecolex.org/}{\color{blue}https://www.ecolex.org}} PINE—operated by the OECD—focuses on environmental policy instruments but does not contain legal documents to support its statistics and lacks information on many key countries.\footnote{\href{http://oe.cd/pine/}{\color{blue}http://oe.cd/pine}} For example, a search for incentive-based schemes in PINE yields no results for Mexico even though Mexico has several high-profile restoration incentive programs such as Sembrando Vida. 

\subsubsection{Overlapping sectors and jurisdictions}
\label{challenge3}
 To effectively analyze environmental policies, one must consider the effects of competing and complementary policies in sectors like agriculture, mining, and tourism. For example, restoration is regulated not only by departments of natural resources but also by economic, forestry, and agricultural government bodies\cite{Brandt2019TextMP}. Fig \ref{fig:overlappingsectors} in Appendix \ref{appendixA} shows an example of the diversity of government bodies in Mexico that have issued legislation containing restoration terms (as described in section \ref{topicfiltering}) and how they relate to the goal of restoration. Moreover, since environmental concerns exist at global, regional, and local levels\cite{oryan-2016}, it is necessary to understand the actors involved in policy implementation at both the federal and sub-federal jurisdictions, which include states and municipalities.

We believe that artificial intelligence (AI) can help us tackle sustainability challenges, including the three challenges mentioned in this section. To this end, we will first consider the current state of the market for NLP solutions in text analytics, and evaluate whether our proposal is already covered by any of the major providers.

\subsection{State of the market for text analytics solutions}

As we fear a decoupling between the research community and progress made by commercial solution providers, we want to take into account the tools that are commercially available which could be used for policy analysis. Many enterprises are effectively leveraging their data, processes, and applications to derive insights from unstructured text-based data. However, their efforts are still siloed. Enterprises often use separate platforms for document capture, categorization \& classification, and robotic process automation (RPA), even though these are all based on similar text mining and text analytics technologies\cite{ForresterDF}. AI can be knowledge-based (based on domain ontologies and language-specific rules) or machine learning (ML)-based. While each approach has its limitations, a hybrid ML- and knowledge-based approach produces the most accurate and timely results, which is reflected in the current trend for text analytics products\cite{ForresterPF}.

After looking at the the main solution providers, which range from large generalist SaaS providers like IBM and Google to smaller specialist text analytics providers like Clarabridge and Micro Focus, we found that our proposal of a specific pipeline for policy analysis fills a niche which has not yet been covered by current market players. We will build this proposal around the use-case of policy in the field of forest and landscape restoration.

\subsection{Landscape restoration as a use-case}

As we enter the UN Decade on Ecosystem Restoration\cite{UNDecade}, making significant progress in the restoration of fragile ecosystems is of utmost importance. The scaling of economic and financial incentives for forest and landscape restoration presents one way to tackle the goals and commitments associated with the UN Decade. In recent years, economic and financial incentive schemes such as payments for ecosystem services (PES) have been some of the most prominent mechanisms by which countries pursue compliance with international agreements like REDD+\footnote{\href{https://redd.unfccc.int/}{\color{blue}https://redd.unfccc.int/}} that aim to offset emissions from deforestation\cite{SIMS20178}.

To drive restoration impact, it is first necessary to understand the range and nature of public incentive schemes that already exist. The design and implementation methods of these incentives vary across countries, regions, and landscapes. One way to understand the variance of incentives is to consider the different types of instruments which are utilized in different policy contexts. Policy instruments—or mechanisms through which economic and financial incentives for restoration are implemented—include tax benefits, direct payments, fines, loans, supplies, and technical assistance (Table \ref{policies} in Appendix \ref{appendixA}). These are the incentive instruments that we focus on identifying in policy documents.

\subsection{Knowledge management system for environmental policy analysis}
\begin{figure*}[h!]
\centering
\includegraphics[scale = 0.48]{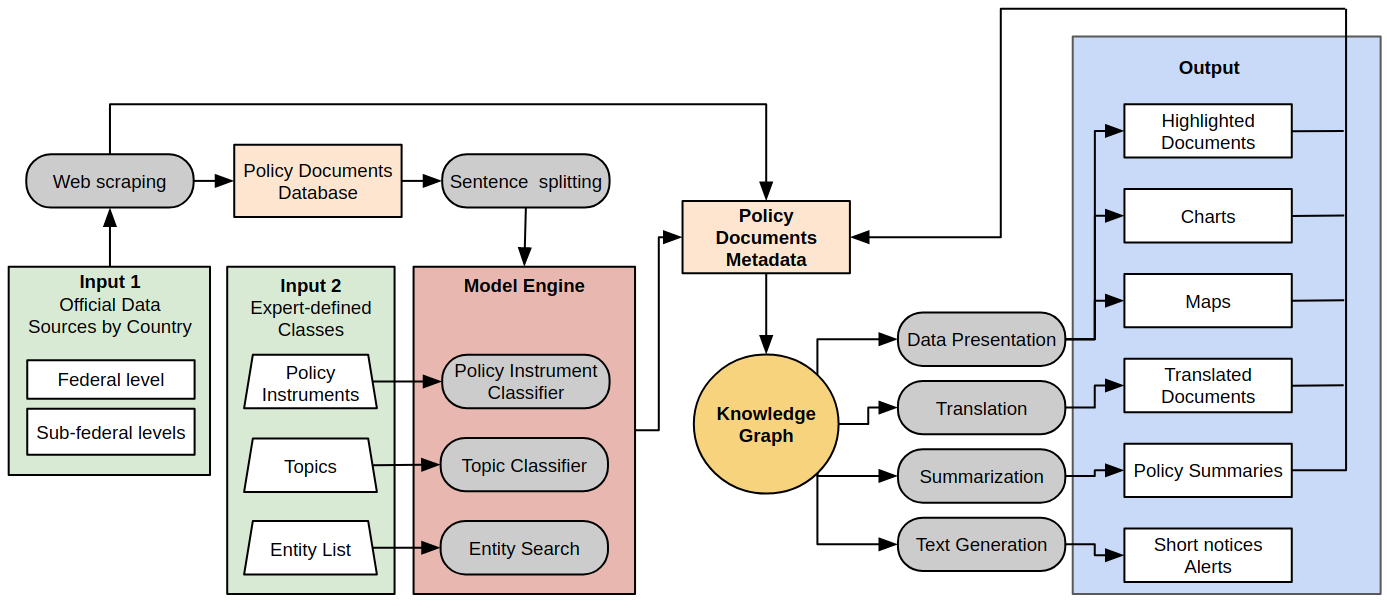}
\caption{Knowledge Management System combining different NLP methodologies to construct an end-to-end pipeline}
\label{KMS}
\end{figure*}

In this paper, we propose a scalable, comprehensive, end-to-end pipeline to identify, retrieve, process, classify, and display information from environmental legislation (Figure \ref{KMS}). Each of the components of our pipeline is built to address a specific policy analysis challenge while offering high performance and resource efficiency. Our ultimate goal is to build a knowledge graph based on the outputs from the framework components, where one would be able to identify and establish relationships among actors, rules, and practices, as these insights are essential for policy analysis.

Having centralized information is an important first step towards monitoring progress in the UN Decade on Restoration and identifying and bench-marking best practices. However, it is not the only step needed for better policy analysis as currently there are very few examples of visualizations.\footnote{{\href{https://www.eea.europa.eu/data-and-maps/daviz/stages-in-the-adaptation-policy-1\#tab-chart\_1}{\color{blue}https://www.eea.europa.eu/data-and-maps/daviz/stages-in-the-adaptation-policy-1\#tab-chart\_1}}} A database with rich metadata would allow for building interfaces and dashboards that highlight regional differences in policies (e.g., geo maps), the frequency of instrument use (e.g., heatmaps), overlap of topics (e.g., correlation plots), and progress towards restoration goals (e.g., time-series analysis).   
 
Our framework is flexible and extensible, both from a technical and a policy perspective. We designed the system in a platform-agnostic manner, meaning that one could use any existing open-source or commercial tool for each component in the pipeline. Given the flexibility of existing multilingual language models\footnote{{\href{https://github.com/google-research/bert/blob/master/multilingual.md\#list-of-languages}{\color{blue}	https://github.com/google-research/bert/blob/master/multilingual.md\#list-of-languages}}} in NLP\cite{conneau-etal-2020-unsupervised}\cite{wolf-etal-2020-transformers}, our approach can be extended to as many languages as the models have been trained for. Furthermore, while we focus on financial incentives and forest landscape restoration policy, our work can be extended to any type of environmental policy. For example, our pipeline could be adapted to track the progress of agreements like the UN Framework Convention on Climate Change\cite{UNFramework}. Additionally, our system allows for automatic updates to include new sources, new legislation, and changes in policy categorization as they emerge.

We believe that our approach would:
\begin{itemize}
\item Reduce the time taken by policy analysts to extract relevant information from documents from weeks to seconds.
\item Serve as a central information source which guarantees full coverage of all policy documents that are electronically available per country, currently categorized by financial incentives.
\item Provide insight on how policies are implemented across regions and how actors are involved in the implementation.
\item Improve institutional transparency, promote alignment within governments, empower stakeholders, and inform future policy developments.
\end{itemize}

Beyond all of these improvements, our proposal is a breakthrough as policy analysts will be able to engage in policy analysis at a global scale.

This paper is organized as follows: In Section \ref{pipeline}, we detail the design of our pipeline and the relevance of each component in the policy information extraction process. In Section \ref{discussion}, we review the state-of-the-art methods for each NLP technology in our pipeline, and discuss the challenges that arise when building the framework. Finally, we conclude by highlighting the importance of proper pipeline design in efficient and effective policy analysis.

\section{Pipeline}
\label{pipeline}
\subsection{Data Collection}
The initial component of our pipeline focuses on collecting raw data, which are the policy documents issued by governments. The diversity and spread of policy sources might seem overwhelming, but there is an underlying principle: no policy is enforced before being passed and subsequently published in a formal source like official newspapers, bulletins, or governmental websites. Therefore, official documents can be retrieved from the predefined official sources for a given nation, state, or geographical region. 

The first component of the pipeline is a crawling tool that scrapes official websites and downloads policy documents and policy metadata. Next, we would need a relational database to store metadata and a cloud facility to store documents. To keep an up-to-date object storage of documents, we propose running the web crawler periodically, either according to a time period pre-defined in the metadata or whenever a change is detected in the official website. 

\subsection{Document Pre-processing}
For a document to be processed computationally, it must be in plain txt file format. Legal texts in electronic form are offered in a diversity of formats. In some cases like the Chilean official website or the US Federal Register, legal texts can be downloaded in different formats, one being the plain text format. In other cases, the documents can only be downloaded in HTML (e.g, Mexico) or PDF (e.g, Peru, El Salvador, Guatemala, and India). There are two main types of PDF documents: either the text is embedded in the PDF markup system or the text belongs to scanned images of the actual document.

The document pre-processing step needs to be adapted to this diversity of situations. For the documents in HTML format, either HTML selectors can be used directly in the scraping tool or the  BeautifulSoup4\footnote{\href{https://pypi.org/project/beautifulsoup4/}{\color{blue}{https://pypi.org/project/beautifulsoup4/}}} library can be integrated within the scraping tool so that the output of the crawling process is a file in txt format. For the transformation of the pure PDF documents into txt files, there are a few python libraries such as pdfMiner\footnote{\href{https://pypi.org/project/pdfminer/}{\color{blue}{https://pypi.org/project/pdfminer/}}}, but these tools are not useful to treat scanned images. To tackle this challenge, we recommend AWS Textract\footnote{\href{https://aws.amazon.com/textract/}{\color{blue}{https://aws.amazon.com/textract/}}}, which is a machine learning service that extracts text from scanned documents.

Within a document, the basic unit of analysis should be the sentence, as documents can contain multiple policies and they are generally written as sentences. Furthermore, the NLP models we propose to use have a sequence length limit, which makes it difficult to feed entire documents as inputs to the models. Thus, we recommend using a combination of Python's NLTK library\footnote{\href{https://www.nltk.org/}{\color{blue}{https://www.nltk.org/}}} and custom functions per document language to split the document into sentences after removing unwanted tags, acronyms, extra spaces, and accent marks.

\subsection{Assisted labeling}
Quality data labeling is a key factor in supervised model performance. There are a few techniques that can be applied to optimize the amount of resources allocated to this process. For our use-case, labeling sentences manually as being incentives or containing a certain incentive instrument is a daunting task because there might be just a few of these sentences in a document that is several pages long. Therefore, we propose an assisted labeling methodology as a special case of active learning. Using out-of-the-box, pre-trained language models to query a policy database would allow the analyst to focus on the labeling of a subset of sentences that are semantically related to incentives, as opposed to all the sentences in a document.

\subsubsection{Query design}
Query design is a sensitive step that requires careful human involvement. We outline a solution based on our use-case, designed in collaboration with policy experts. First, a set of sentences stating incentive instruments in real policy documents must be built for each language. These sentences may need some trimming to remove any element which could divert the model from focusing on the incentive itself. In order to be able to keep the idiomatic nuances of different dialects, sentences can be derived from different countries. For example, by picking a model sentence from each of five different LATAM countries for each of the six incentive classes, we can build a list containing 30 different queries.

\subsubsection{Labeled datasets}
To build training datasets, we propose a collaboration between NLP models and policy experts. First, the embeddings of each query and all the sentences in the database are computed using pre-trained Sentence-BERT (SBERT)\cite{reimers2019sentencebert}. The cosine similarity between each sentence and each query is calculated and the sentences ranked by similarity score. After duplicates are removed, a dataset of pre-labeled sentences for each incentive class is ready for further processing. We use the term ‘pre-labeled sentences' because the sentences are assigned a label based on a similarity score to the queries representing a given class. Finally, the label of pre-labeled sentences is confirmed or rejected by an analyst. The compiled labeled sentences will be used as a model training set.

\subsection{NLP methods}
\label{NLP methods}
Our knowledge management system relies heavily on NLP methods to extract relevant information that will later be compiled in a knowledge graph for retrieval by policy analysts. In these subsections we describe the NLP tasks that help us build the knowledge graph.
\subsubsection{Model Engine} One of the nominal breakthroughs in the field of NLP was the release of BERT\cite{devlin2019bert}, a model that outperformed all other previous models in many tasks. In this work, BERT plays a central role in the NLP pipeline, which we call the Model Engine. Conceptually, this engine is a cluster of BERT-based models that are fine-tuned for different tasks. However, even though we propose to use Transformers\cite{Vaswani2017} as a baseline for most of our NLP tasks, we are aware of current work\cite{mlp} that is attempting to improve the cost and performance of BERT-based models. Therefore, we acknowledge that the Transformers in our Model Engine could be replaced by other architectures with better multi-task precision/recall performance, as well as computational cost.  
 
\subsubsection{Incentive and topic classification} In the use-case that we employ for pipeline design and development, our main goal is to find economic and financial incentives in policy documents and link them to the appropriate land-use types (e.g, forest or agriculture). As explained above, incentives can be classified into six different categories: tax benefits, direct payments, fines, loans, supplies, and technical assistance. Thus, the core model engine will include (1) a fine-tuned BERT-based model for identifying the presence of incentives in a sentence and (2) a fine-tuned model for incentive classification. In complex legal documents, several incentives might be applied to different goals. As our text unit is the sentence, we seldom have the information about the target of a given incentive in the sentence where the incentive is stated. Hence, we need to solve the problem of assigning each incentive to a target topic—for example, a land-use type. Brandt\cite{Brandt2019TextMP} suggested using semantic search at the sentence level to label sentences as referencing a topic or not. Therefore, (3) a fine-tuned BERT model for topic classification is also included in this pipeline, where each sentence gets classified into one of many of pre-defined topics. 

At this point, sentences are tagged by incentive and by topic. In the most simple case, if there is only one sentence containing incentives or only one sentence belonging to a topic, there is no risk of mismatch. In more complex documents, different incentives can be matched to different topics. We propose to match incentives with topics by the minimum distance approach. Here, an incentive will be linked to its closest sentence topic, the distance measure being the number of sentences between two tagged sentences. While this method is efficient and straight-forward, it may lead to some conflicts where the model identifies more than one topic per sentence, in which case the results should be verified by a policy analyst.

\subsubsection{Named Entity Recognition and Relationship Extraction}
Named Entity Recognition (NER) is the process of identifying named entities in unstructured text and classifying them into predefined semantic categories such as person, organization, and location. Relationship extraction (RE) simply identifies the relationships between two entities. For the environmental policy domain, these two NLP tasks provide insights to the policy analyst about the relationship among actors, program implementation details, and conflicting or competing dynamics among government agencies. For example, named entities that can be extracted from a document are a governmental entity like the Mexican Secretaria of Bienestar, a program implementing a policy like "Sembrando Vida", and the people/regions targeted by such program like land owners in specific rural areas. The relationships between program administrator, policy implementation, and policy target would be extracted from the document in an automated manner as well. NER/RE models would be essential components of the Model Engine, and we describe the methods proposed for these tasks in section \ref{nerre}.

\subsubsection{Summarization}
One way to increase transparency and accessibility, especially in the context of the decentralized nature of restoration legislation is through providing summaries. The availability of summaries varies across countries. For example, US Federal Bills come with human-written summaries from the Congressional Research Service\footnote{\href{https://www.govtrack.us/}{\color{blue}{https://www.govtrack.us/}}}, and major EU legal acts are summarized by Eur-Lex\footnote{\href{https://eur-lex.europa.eu/}{\color{blue}{https://eur-lex.europa.eu/}}} for the general audience in the languages of the member states. In other countries and regions, the availability of summaries is inconsistent. As an integral part of the proposed pipeline, summarization can not only provide quick and relevant information, but also enrich the document database by adding short searchable summaries.

\begin{figure}[h]
\includegraphics[scale=0.30]{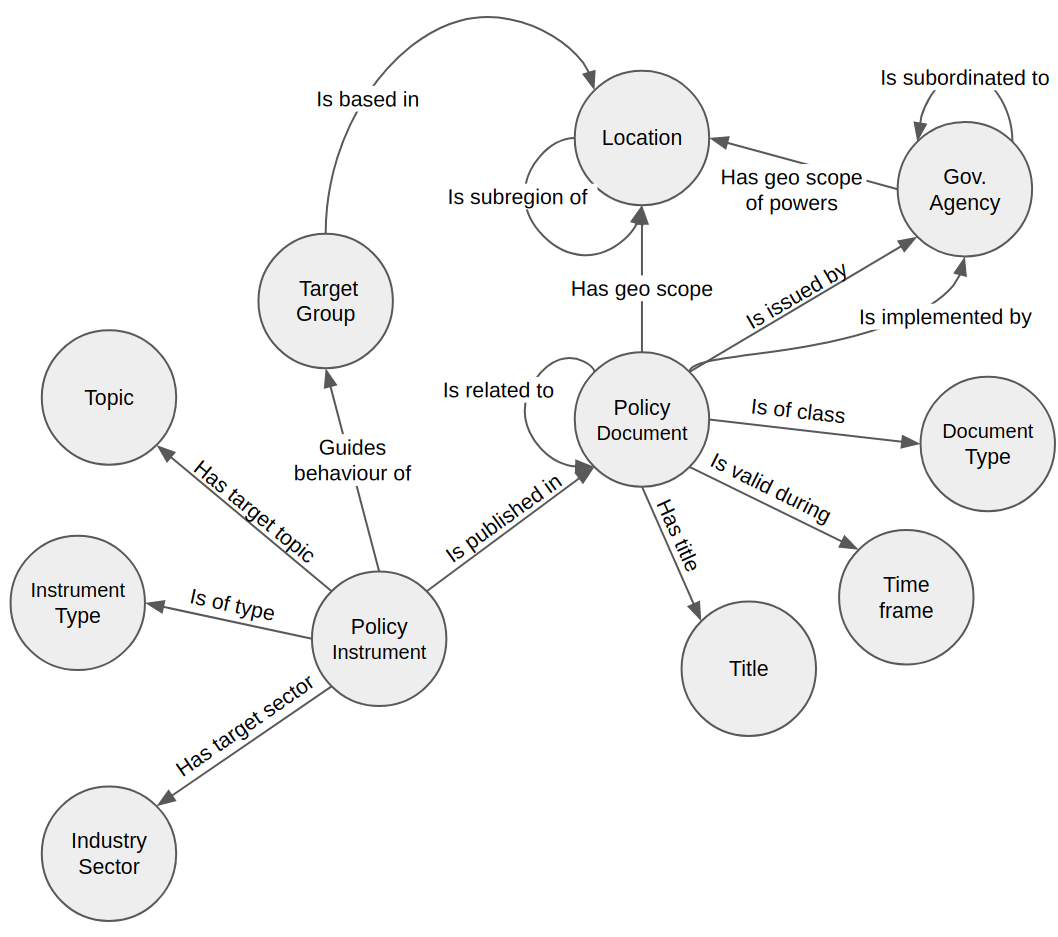}
\caption{Knowledge Graph Ontology outlining the relationships among relevant policy entities}
\label{fig:kgontology}
\end{figure}

\subsection{Knowledge Graph}

Given the number of concepts that can be drawn from policy documents, such as government entities, program names, incentive instruments, topics, and the different relationships that can exist between each of these concepts, we choose to store them in a knowledge graph (KG). The first step to build a KG is to define an ontology, which is a structured semantic data model for a specific domain (in our case, environmental policy documents) that defines the types of concepts that we can store and the properties that can be used to describe them. Specifically, we will define classes, their attributes, and their relationships. Wang et al.\cite{Wang2019} proposed a KG ontology that was meant to tackle the general relationships between entities in a policy document. In our proposal, we tailor the ontology to store concepts and relationships encountered in restoration policy analysis. 

As an example, we will look at a policy sentence from the Wisconsin Forestry Grant and State Aid Administration \cite{Wisconsin}. 

\begin{itemize}
\item \textit{Example:} NR 47.10 “The purpose of this subchapter is to establish procedures and standards for the administration of the stewardship incentive program (SIP) as authorized under the act, for the purpose of encouraging private forest land-owners to manage their lands in a manner that benefits all the resources in their forest.”
\item \textit{Disambiguation:} “stewardship incentive program (SIP)” \textit{[program name]} “private forest land-owners” \textit{[target of program]}, forest\textit{ [land-use type]}, 
\end{itemize}

The decomposition of the sample policies can be abstracted into a format similar to that of the Resource Description Framework\footnote{\href{https://www.w3.org/TR/rdf-concepts/}{\color{blue}{https://www.w3.org/TR/rdf-concepts/}}}: a triple of (subject, predicate, object). In the example above, a valid triple could be \textit{(“Stewardship Incentive Program (SIP)”, “targets”, “private forest land-owners”)}. Other valid triples derived from information in document metadata, topic analysis, or keywords would be \textit{(“The Stewardship Incentive Program (SIP)”, “has scope”, “federal”)} or \textit{(“The Stewardship Incentive Program (SIP)”, “has objective”, “reforestation and afforestation”)}. After manually analyzing some sample policy documents, we abstract our RDF triples with the ontology shown in Figure \ref{fig:kgontology}. Here, we identified \textit{Policy, Policy Instrument, Target Group, Location and Government Agency} as the main classes, which can be extracted from our NER component, as well as our sentence splitter for policies. Attributes such as \textit{Topic and Policy Instrument type} can be extracted using our classification component, and the rest will be extracted using custom functions.

\noindent
\begin{table*}
\renewcommand{\arraystretch}{1.3}%
\begin{tabular}{ | m{3.7cm} | m{5cm} | m{3.7cm} |  m{3.9cm} | }
\hline
\multicolumn{1}{|c|}{\textbf{Challenge}} & \multicolumn{1}{c|}{\textbf{Problems within Challenge}} & \multicolumn{1}{c|}{\textbf{Process (pipeline part)}} & \multicolumn{1}{c|}{\textbf{Output}} \tabularnewline
\hline
\multirow{4}{*}{Time- and resource-intensity} & Lengthy documents, inconsistent language & Incentive classification & Incentive identification \tabularnewline \cline{2-4} 
 & Lengthy documents, text complexity & Summarization & Document summaries \tabularnewline \cline{2-4} 
 & Language barrier & Translation & Text/summaries in English \tabularnewline \cline{2-4} 
 & Complex structure & Document Outline & Document structure, Relevant parts of document\tabularnewline \hline
\multirow{3}{*}{Information accessibility} & Multiple documents, decentralized sources, remit overlap & Scraping with Topic Filtering & Document Collection \tabularnewline \cline{2-4} 
 & Decentralized sources, lack of metadata, dynamic information & Database & Document metadata \tabularnewline \cline{2-4} 
 & Format inconsistencies, difficulties in processing & Pre-processing , Standardization & Searchable documents in standardized format \tabularnewline \hline
\multirow{3}{*}{\begin{tabular}[c]{@{}l@{}} Overlapping sectors \tabularnewline and jurisdictions\end{tabular}} & Gaps and dependencies between documents, coverage & NER (on metadata and on text) & Identifies actors, policies, geo, Linked entities/policies map\tabularnewline \cline{2-4} 
 & Topic overlap in lengthy documents, text complexity & Topic modeling Knowledge graph & Topic identification/overlap \tabularnewline \cline{2-4} 
 & Difficulty in connecting multiple concepts, entities, and topics & Knowledge graph & Relationship between concepts, entities, topics \tabularnewline \hline
\end{tabular}
\caption{Policy Analysis Challenges and Solutions using the proposed NLP pipeline}
\label{challengesolution}
\end{table*}

\section{Discussion}
\label{discussion}
The three challenges of policy analysis (section \ref{policyanalysischallenges}) can be broken down into specific problems caused by the nature of legal documents and legislation practices (Table \ref{challengesolution}). These problems can be tackled by introducing NLP solutions whose output would help analysts to speed up the policy analysis process (e.g. through summaries), centralise information (e.g. through a database), and understand the interplay between the various actors (e.g. through entity graphs). This section presents a discussion about the methodology, latest developments and limitations, and challenges for the proposed solutions.  
\subsection{Data collection}
Automated data collection is the solution we propose to tackle the challenges we describe in sections \ref{challenge2} and \ref{challenge3}. Accessing public repositories of official documents through web crawling and scraping ensures an updated and comprehensive retrieval of documents at different administration levels. Here we discuss some more specific challenges of the automation of the data collection process.

\subsubsection{Data comprehensiveness}
The goal of any country’s legal information system should be to achieve a threshold degree of transparency, understood as having the documents that contain a country’s laws accessible both to those governing and to those being governed. Moreover, analysis of the use of information technology to achieve sustainable transparency starts with two criteria: first, the database containing a country’s laws must be comprehensive, accurate, and timely; and second, user access to the database must be convenient, easy, and reliable \cite{Yates2020}. However, the level of ‘e-government' sophistication varies across countries. The United Nations ranks countries according to the E-Government Development Index (EGDI) with indices ranging from “low” to “very high” based on factors such as online services and open government data. According to the UN’s 2020 E-Government Survey, 58\% of European countries have very high EGDI levels, while only 26\% of Asia, 12\% of the Americas, and 4\% of Oceania belong to the very high EGDI group \cite{UNDESA}.

In our experience, there are e-government sophistication differences both among countries and within them. For example, while the U.S and Mexican federal governments provide APIs to retrieve official documents from their websites, the same accessibility standard is not guaranteed at the state level. In other countries like India, there is a central interface (e.g., the India Code) from which legislation at both the federal and state level can be accessed. This is and will be a limiting factor of any automated pipeline for policy analysis that cannot be addressed using technology, and it has to be clearly stated when the results of the analysis are presented.

\subsubsection{Topic filtering}
\label{topicfiltering}
Public electronic records of legislation contain all types of legal documents and can reference any topic of the legislative domain—from foreign policy to telecommunications. At some point in the pipeline, some kind of filtering should be applied. The filtering could be applied when the ML models are used to retrieve information from documents, but this would impose a burden on the amount of information to be first stored and later on processed. On the other hand, some e-portals are only accessible through thematic filters, so the user has to select a keyword to get a list of results. Thus, a filtering step during web crawling might be suitable and, in some cases, compulsory.
We advocate for a two-step filtering process. First, a gross filter to be applied during the crawling process to select only legal documents which broadly fall under the scope of environmentally sensitive areas. A fine-grained filter will later be applied after the documents are processed by the model engine to select the documents containing the desired information. 

For the filtering step during the crawling process, our proposal consists of applying keyword filters, in-keyword and out-keyword filters. The in-keyword filter is a comprehensive list of keywords related to landscape restoration. However, since keywords are polysemantic, some of the retrieved information might be out of scope. In this case, the out-keyword filter would refine the results from the first filter. This approach is simple and fast, but difficult to scale for two reasons. First, lists of keywords must be created for each language. Second, while the in-keywords list is quite universal as it depends on the topic, the out-keyword list has many dependencies on the specific legal vocabulary of each country. Taking into account that the most important metric to consider in the development of these filters is recall, the use of spam-filtering techniques could help to avoid missing important documents and to develop a more scalable tool in the future\cite{garg2021}.

For fine-grained filtering, we choose a simple approach suggested by Brandt\cite{Brandt2019TextMP}. After splitting a document into sentences, we classify the sentences by whether they belong to one of the topics of interest. The document will be assigned to as many topics as sentences are assigned to different topics.

\subsection{Text pre-processing}
The length and the structure of documents relevant to policy analysis are diverse. Some notices might have one or two paragraphs under the title, while other laws might have more than one hundred pages with the text structured in several sections and subsections.
The main advantage of retaining document structure is that some basic information is always kept in certain sections. For example, references to other legal texts are usually presented in specific sections at the beginning of the document. If the structure is retained during pre-processing, the searches that will be made downstream, like searching economic incentives in our use-case, can be targeted to specific sections, optimizing time and computer power. However, keeping document structure is not straightforward, as there is no standard document scaffold and every issuing institution applies its own rules. This means that the pre-processing step must be tailored to every type of document. Finding a way to generalize the process of retaining document structure remains a challenge that will benefit document processing, and the unsupervised method by Aliyu et al\cite{Aliyu2020}. may well be a first step in this direction.

\subsection{Classification}

The classification component of the NLP pipeline deals with identifying and retrieving relevant information in policy documents, as explained in section \ref{NLP methods}. This pipeline component focuses on reducing the time and resource requirements for information retrieval and making useful information accessible to stakeholders, as stated in sections \ref{challenge1} and \ref{challenge2}, respectively. Once the analyst's goals are set and the model is trained, consistently finding the information in numerous complex documents will take a matter of seconds.

Sentence classification has been vigorously studied and developed in the past few years. Some of the best performing models include SciBERT, which is pre-trained on a large multi-domain corpus of scientific publications in an unsupervised way\cite{scibert2019}; and FE-MLM + Span which improves self-supervised pre-training using a fully-explored Masked Language Model\cite{mlm2020}. Among different techniques, the best performing models use some variation of a Transformer, largely because self-attention is a theoretically and practically stronger technique to process text than rule-based and traditional sequential models are\cite{shervin2020}.

Classification models can be evaluated based on their precision and recall performance. To account for class imbalance, the $F_1$ score (harmonic mean of precision and recall) is popularly used. For policy analysis, it would be better to have a higher recall than precision score, i.e., the cost of a false negative is more concerning than that of a false positive. The trade-off we deal with is getting all the relevant information with some noise as opposed to having a small amount of noise but losing relevant information. For example, a sentence discussing tax deductions incorrectly missed by the model is more detrimental to a policy analyst's work than having a small number of false positives in the tax deduction predictions. Maximizing recall would assure a policy analyst that all (or as many as possible) tax deduction instances are classified as such, at the affordable cost of some rogue sentences classified incorrectly as well.

There are two problems which require a classification model, assisted labeling and sentence classification. For assisted labeling we propose using SBERT which excels at computing sentence embeddings\cite{reimers2019sentencebert}. In assisted labeling, SBERT is more efficient in run-time as opposed to other notable methods such as Cross-Encoders when embeddings for large datasets are computed, as we need to compare a large number of sentence pairs (queries and policies)\cite{reimers2019sentencebert}. For sentence classification both SBERT and Cross-Encoders perform well and can be used according to the specific problem they are being applied in. For sentence classification inference with SBERT, a random forest classifier can be used to assign a category to a given sentence using the learned embeddings from the SBERT model\cite{ourpaper} as input. Pre-trained SBERT models can be found on Huggingface's official repository.\footnote{\href{https://huggingface.co/sentence-transformers}{\color{blue}{huggingface.co/sentence-transformers}}}

\subsection{Named Entity Recognition and Relation Extraction}
\label{nerre}
NER and RE are the NLP tasks we propose to address the challenges from section \ref{challenge3}. In most NER model architectures, there are three main components: input representations, encoder, and decoder. Generally, significant performance improvements are achieved when the input representations come from pre-trained language model word embeddings. In the work by Wang et al.\cite{Wang2019}, the entity recognition component was based on using a CNN for character encodings as input, a BiLSTM for encoding, and a conditional random fields (CRF) model for decoding\cite{Ma2016}. The combination of BiLSTM+CRF is the most common architecture for NER using deep learning, and it is used in one of the top-performing methods—ACE\cite{Wang2021automated}—in benchmark datasets. 

However, some key points raised by Li et al.\cite{Li2020} are that Transformers are more efficient and more effective than LSTMs as encoders if they are pre-trained on a large corpus. Similarly, CRF models as decoders can be computationally expensive when the number of entity types is large, and they do not always lead to better performance compared with a softmax layer combined with contextualized language model embeddings (i.e., BERT and ELMo). For instance, both the state of the art for the OntoText5.0 English dataset\cite{OntoNotes} and the second top-performer in the CoNLL03 English benchmark\cite{sang2003} use a BERT-based model. 

Relationship extraction methodologies can act at the sentence or document level. Wang et al.\cite{Wang2019} proposed to use a CNN-based method\cite{Zeng2015} evaluated in the NYT+Freebase dataset\cite{Riedel2010}, a corpus with 430 different relation-types. Yet NLP models for RE have since then improved dramatically. To encode representations for both entities and their attributes or related context within a corpus, the state-of-the-art methods for the dataset from Riedel et al.\cite{Riedel2010} use a combination of Graph Neural Networks, self-attention and rules from a knowledge graph. Other methods for datasets with fewer and more general relation types (\cite{zhang2017tacred},\cite{Hendrickx2010}) leverage a pre-trained transformer-based model such as RoBERTa\cite{Zhao2021},\cite{ZhouChen2021}. The embeddings created at this point get passed to a classifier that predicts the correct relation between entities\cite{Bastos2021},\cite{Nadgeri2021}.

For policy-analysis, it is important to note that (1) there are not many possible relationships between entities, (2) relations are not always explicitly stated in sentences (e.g.,"the state forester" does not mention that it is Winsconsin's state forester), (3) we need methods that are able to generalize to previously unseen entities, as we may not always have enough data available per country and (4) we need methods that can be easily extended to many languages. Given these constraints and the success of transformers in creating good text representations, we propose to use state-of-the art transformer models that are openly available and ready to be used out-of-the-box. For both NER and RE, we recommend using HuggingFace’s LUKE\cite{Yamada2020} implementation\footnote{\href{https://github.com/studio-ousia/luke\#news}{\color{blue}{https://github.com/studio-ousia/luke\#news}}}, the pre-trained NER model from ACE\footnote{\href{https://github.com/Alibaba-NLP/ACE}{\color{blue}{https://github.com/Alibaba-NLP/ACE}}}, or Zhou and Chen's RE model implementation\cite{ZhouChen2021}.\footnote{\href{https://github.com/wzhouad/RE\_improved\_baseline}{\color{blue}{https://github.com/wzhouad/RE\_improved\_baseline}}} Finally, we can also create rules to automatically extract some of the term definitions that are traditionally included at the beginning of policy documents.

\subsection{Summarization}

Automatic summarization methods have been studied on a variety of specific domains\cite{zhang2020pegasus}, including legislation texts\cite{zhang2020pegasus},\cite{Eidelman_2019}. While unsupervised extractive summarizing methods (SumBasic, LSA, TextRank) retain veracity by literally reproducing key phrases from the input text, they may have problems with lengthy documents rich in generic legal text. A good legislation summary should highlight major actions, skip background content, and typically contain more action verbs and entity names\cite{Eidelman_2019}, effectively answering the ‘what' and 'who' questions. Abstractive summarization methods first build an internal semantic representation and then use natural language generation techniques to create a summary\cite{GUDIVADA2015203} but are found to not always represent content faithfully\cite{maynez2020faithfulness}. However, techniques like BillSum’s BERT-based methods\cite{Eidelman_2019} and Google’s Pegasus gap-sentences generating model\cite{zhang2020pegasus} perform well in legislation summarization tasks and produce good ROUGE scores\cite{lin-2004-rouge} on unseen sets of data, thus suggesting that developed summarization methods could be used for legislatures without human written summaries. In particular, BillSum’s DOC+SUM method\cite{Eidelman_2019} outperforms the rest by combining linguistic information with document structure information (e.g., place of sentence, keyword relevance).

Since assessing a summary’s quality is subjective, automatic summarization is an area where human input is embedded. The ROUGE metric for summary quality uses comparison to human-generated summaries. The authors in \cite{zhang2020pegasus},\cite{Eidelman_2019} and \cite{NEURIPS2020_1f89885d} use human-generated summaries as a base and benchmark. Openai’s GPT-3\cite{NEURIPS2020_1f89885d} uses human feedback by training a reward model via supervised learning to predict which summaries humans will prefer. Moreover, the capabilities of GPT-3 can be used to tackle summarization and translation simultaneously, which is possible depending on the prompt and parameters used.\footnote{\href{https://beta.openai.com}{\color{blue}{https://beta.openai.com}}}

Document text volume represents a limitation to successful legislation summarization. BillSum\cite{Eidelman_2019} only deals with mid-length legislation, whereas GPT-3 puts a limit on processed tokens, making it difficult to process original documents without splitting them up. However, in combination with appropriate pre-processing, document structure, and entities knowledge, it may be possible to achieve focused summaries with even higher relevance to a specified topic.

\subsection{Responsible ML principles}

Concerns about bias and unwanted consequences in applying ML have resulted in multiple models for governance\cite{Gasser_2017} and practical checklists.\footnote{\href{https://digital-strategy.ec.europa.eu/en/library/assessment-list-trustworthy-artificial-intelligence-altai-self-assessment}{\color{blue}{https://digital-strategy.ec.europa.eu/en/library/assessment-list-trustworthy-artificial-intelligence-altai-self-assessment}}}\textsuperscript{,}\footnote{\href{https://ethical.institute/principles.html}{\color{blue}{https://ethical.inste/principles.html}}} Although the content varies, there is broad consensus on the importance of human interaction, avoidance of bias, and general principles such as privacy and security. In the domain of restoration, linking ML models to traditional ecosystem knowledge would align incentives between human and algorithmic actors\cite{rakova2020leveraging}. In the proposed pipeline, human-in-the-loop interaction points would include expert knowledge in the process. For example, expert labels would be used for data augmentation, improving the capability of recognizing the relevance of sentences to queries. To reduce subjectivity, multiple raters would rate the query relevance and inter-rater reliability would be calculated\cite{Green97}. Additionally, subject matter experts would regularly evaluate the output of the various steps. To counter the bias of English-centric NLP models\cite{bender-2019}, performance on non-English documents would be evaluated too. Explainability and transparency would be achieved by the design of the pipeline which allows tracing of information through the steps, as well as through visualizing output and metrics for clarity. Accuracy metrics should be selected to fit the domain and business goals (e.g., to minimize false negatives when identifying incentives) and the nature of the task (e.g., a set of benchmark documents to assess the quality of scraping). Metrics should be integrated in the major steps, which, alongside the reiterative nature of the ML cycle, enables continuous improvement. Finally, as the pipeline is designed with the cloud in mind, with the possible use of commercial solutions, privacy and security are ensured by the cloud providers’ rigorous security protocols, and built-in recovery and compliance for data protection.

\section{Conclusion}

Effective policy analysis is vital to the success of the Agenda for Sustainable Development. In order to make a long-lasting, positive impact, a holistic Knowledge Management Framework will be useful. The proposed NLP pipeline brings together state-of-the-art methods to retrieve policy documents, process them, extract relevant information and finally, connect these pieces together to deliver insights to the policy analyst. While the pipeline has been designed for a use-case on landscape restoration, it can just as well be fine-tuned to any other environment-focused SDG with domain experts making the necessary adjustments in the data collection, data pre-processing \& labeling, and knowledge graph design. Adapting the latest NLP techniques will help the restoration policy community to combat problems of time \& resource management, information accessibility, and sectoral \& jurisdictional challenges.

\begin{acks}
We thank David Silva and Ramansh Grover from DSSG Solve for their contributions to the project and René Zamora Cristales from WRI for his insights which have been key to understanding end-user needs and taking them into account for the pipeline design.
\end{acks}

\newpage
\theendnotes

\bibliographystyle{ACM-Reference-Format}
\bibliography{references}

\newpage
\onecolumn
\appendix

\setcounter{table}{0}
\setcounter{figure}{0}
\renewcommand{\thetable}{A\arabic{table}}
\renewcommand{\thefigure}{A\arabic{figure}}
\section{Supplementary tables and figures}
\label{appendixA}

\begin{figure*}[h!]
\includegraphics[scale=0.63]{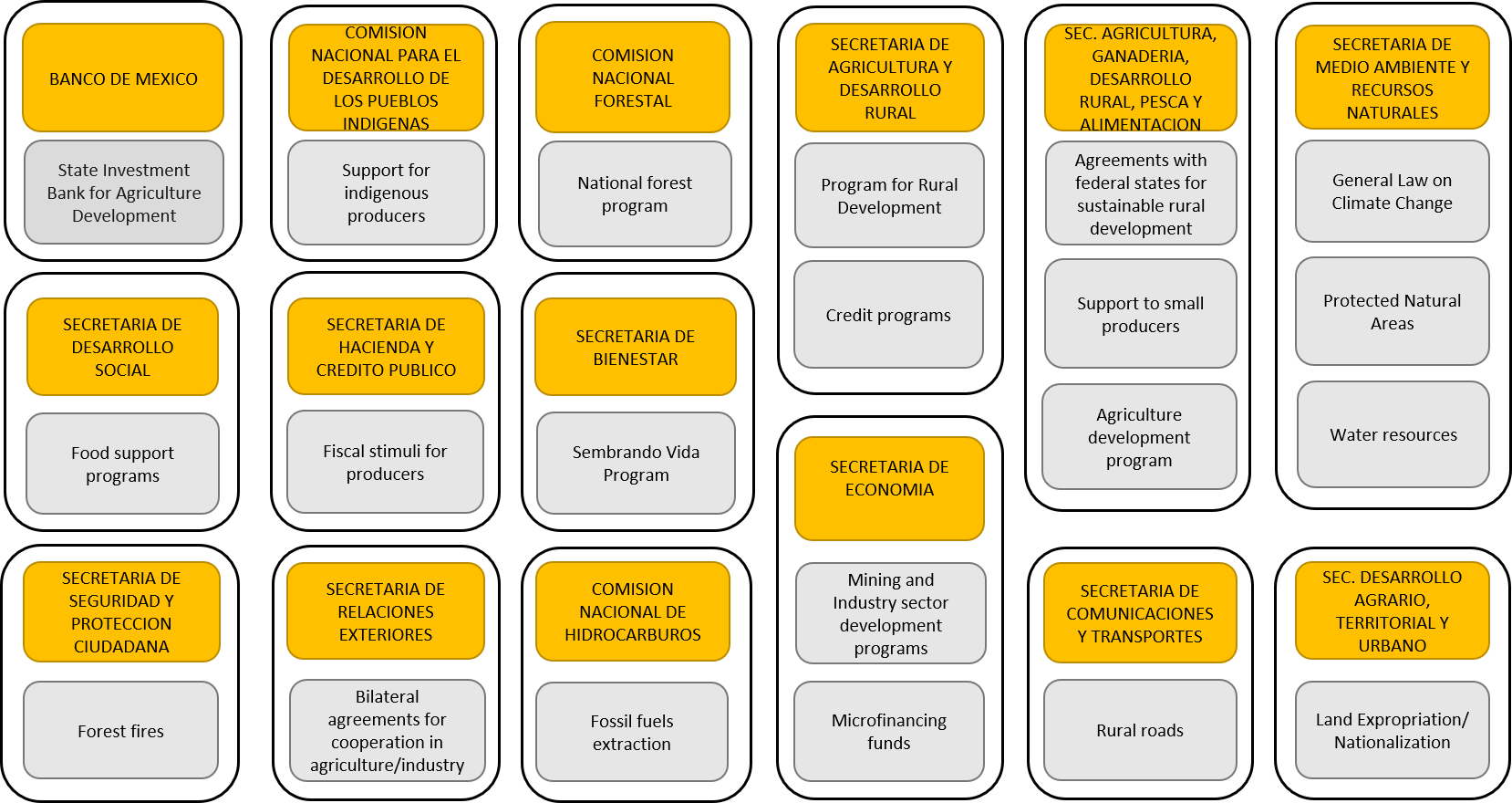}
\caption{Overlapping federal restoration legislation in Mexico – Issuing entities, major programs, instruments and topics}
\label{fig:overlappingsectors}
\end{figure*}

\renewcommand{\arraystretch}{1.6}%
\begin{table*}[htb]
  \centering
    \begin{tabularx}{18cm}{|X|X|X|}
\hline

\multicolumn{1}{|c}{\textbf{Instrument}} & \multicolumn{1}{|c|}{\textbf{Policy}} & \multicolumn{1}{c|}{\textbf{Example}} 

\\ \hline

Tax benefits & Peru: Ley Forestal y de Fauna Silvestre Ley (Forestry and Forest Wildlife Law) & Grants tax deductions to those who conserve land, rehabilitate land, or diversify their forest management to include timber and non-timber trees. Direct payments \& Guatemala: PINPEP (Forestry Incentives for Smallscale Possessors of Forest or Agroforestry Land)

\\ \hline

Direct payments & Guatemala: PINPEP (Forestry Incentives for Smallscale Possessors of Forest or Agroforestry Land)
& Awards direct payments for small landowners who manage agroforestry systems, plantations, and natural forests

\\ \hline
Fines & Chile: Ley Nº 20.653 (Increasing Sanctions for Those Responsible for Forest Fires) 
 & Punishes those who instigate forest fires with monetary fines and, in some cases, jail time
\\ \hline
Loans & United States: California Forest Improvement Program
 & Promotes reforestation and conservation by making loans to cover all or part of landowners’ costs
\\ \hline
Supplies & Mexico: Sembrando Vida (Sowing Life) & Encourages farmers to plant trees among crops (agroforestry) by providing tools, supplies, and plants and establishing community nurseries. Sembrando Vida also awards payments
\\ \hline
Technical assistance & India: National Agroforestry Policy & Provides farmers with access to training centers to learn about agroforestry and silviculture
\\ \hline
\caption{Examples of policy instruments acting as landscape restoration incentives}
\label{policies}

\end{tabularx}
\end{table*}
\end{document}